%% file: main.tex
\documentclass[10pt]{article} %
\usepackage{colm2024_conference}

\input{math_commands.tex}

\usepackage{hyperref}
\definecolor{darkblue}{rgb}{0.00784313725490196, 0, 0.5176470588235295}
\hypersetup{colorlinks=true, citecolor=darkblue, linkcolor=darkblue, urlcolor=darkblue}
\usepackage{enumitem}

\usepackage{graphicx}
\usepackage{url}
\usepackage{booktabs}
\usepackage{multirow}
\usepackage{tcolorbox}
\usepackage{graphicx}
\usepackage{subcaption}
\usepackage{etoolbox}

\let\classAND\AND
\let\AND\relax
\usepackage{algorithmic}

\let\AND\classAND
\AtBeginEnvironment{algorithmic}{\let\AND\algoAND}

\usepackage{algorithm}
\usepackage{algorithmic}

\NewDocumentCommand\emojilogo{}{
        \includegraphics[scale=0.04]{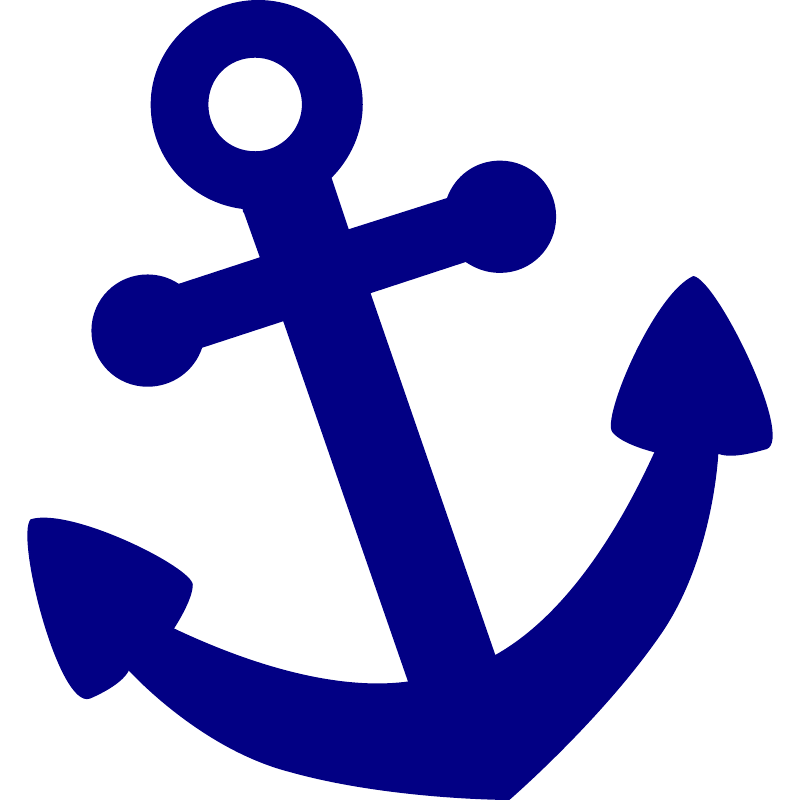}
}

\title{\emojilogo Sailor: Open Language Models for South-East Asia}

\def\month{MM}  %
\def\year{YYYY} %
\def\openreview{\url{https://openreview.net/forum?id=XXXX}} %

\colmfinalcopy
\begin{document}

\author{
Longxu Dou$^{1}$\thanks{The first two authors contributed equally.}\quad Qian Liu$^{1\,*}$\quad Guangtao Zeng$^2$\quad Jia Guo$^1$\quad Jiahui Zhou$^1$ \\
\textbf{Wei Lu$^2$\quad Min Lin$^1$}\\
$^{1}$Sea AI Lab, Singapore \quad\quad
$^{2}$SUTD, Singapore \\
\texttt{\{doulx, liuqian\}@sea.com} \vspace{0.4cm} \\
\textbf{Homepage}: \url{https://sailorllm.github.io} \vspace{0.1cm}\\
\textbf{Model}: \url{https://huggingface.co/sail}
}

\maketitle

\begin{abstract}
We present Sailor, a family of open language models ranging from 0.5B to 7B parameters, tailored for South-East Asian (SEA) languages.
These models are continually pre-trained from Qwen1.5, a great language model for multilingual use cases.
From Qwen1.5, Sailor models accept 200B to 400B tokens, primarily covering the languages of English, Chinese, Vietnamese, Thai, Indonesian, Malay, and Lao.
The training leverages several techniques, including BPE dropout for improving the model robustness, aggressive data cleaning and deduplication, and small proxy models to optimize data mixture.
Experimental results on four typical tasks indicate that Sailor models demonstrate strong performance across different benchmarks, including commonsense reasoning, question answering, reading comprehension and examination.
Embracing the open-source spirit, we share our insights through this report to spark a wider interest in developing large language models for multilingual use cases.
\end{abstract}

\section*{\centering{Takeaway}}

(1) Language models struggle with multiple languages, and continual pre-training presents an opportunity to improve specific language capabilities. (2) Code-switching techniques can be beneficial in multilingual scenarios, improving the ability to handle language mixing. (3) Language models are sensitive to subword segmentation, and techniques like BPE dropout can improve model robustness. (4) Even available high-quality multilingual corpora may require further data deduplication and cleaning. (5) Simulation experiments on smaller models can provide insights into performance trends for large-scale experiments.

\begin{figure}[htbp]
    \centering
    \includegraphics[width=1.0\columnwidth]{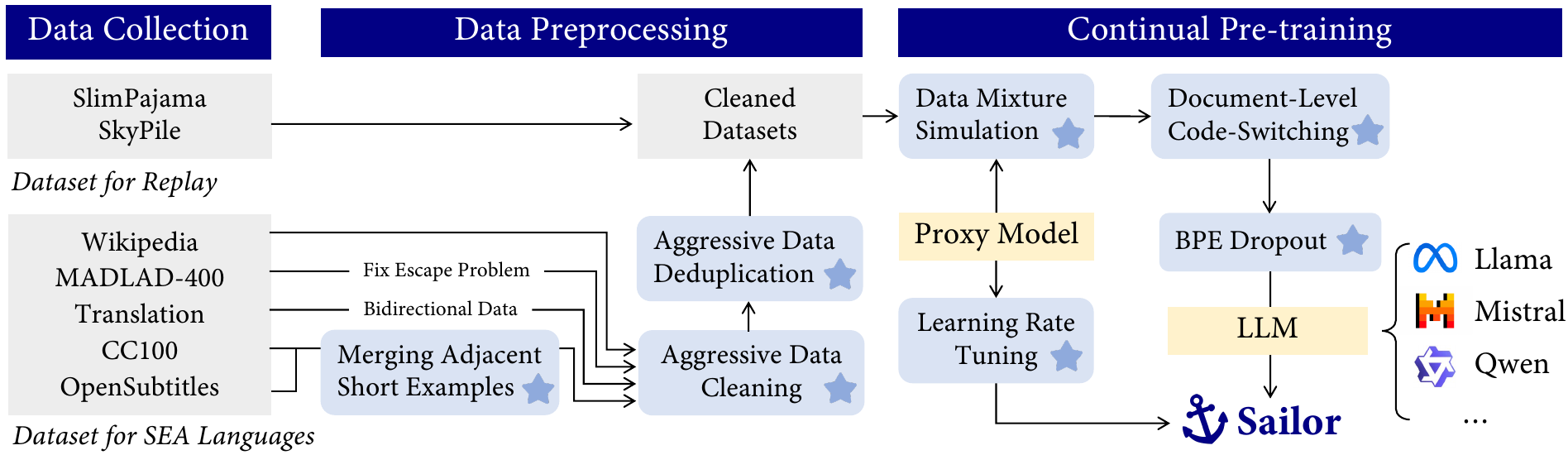}
    \vspace{-1mm}
    \caption{The pipeline of building Sailor, with insights marked by stars.}
    \label{fig:sailor_overview}
\end{figure}

\section{Introduction}

Recent years have witnessed an astonishing surge in the performance of large language models (LLMs), driven by the rapid growth of Internet data~\citep{CommonCrawl2018} and advances in pre-training techniques. The advent of models such as GPT-3~\citep{gpt3paper}, Gemini~\citep{gemini2023}, and Llama
~\citep{llama1paper} has fueled ever-increasing expectations for LLMs across diverse domains, ranging from creative writing~\citep{Weaver2024}, coding~\citep{lozhkov2024starcoder2} and logical reasoning~\citep{agieval2023}.
Developing high-quality models crucially depends on access to a large-scale and high-quality dataset.
The ubiquity of digitized English content has established it as a preeminent source for training LLMs.
Consequently, mainstream LLMs~\citep{llama1paper,ai2024yi,qwen} tend to heavily rely on English datasets. For example, 89.70\% of the training data of Llama-2 is English~\citep{llama2paper}.
However, these English-centric LLMs frequently encounter difficulties in achieving comparable performance across other languages (e.g., Thai). This phenomenon, termed the curse of multilinguality~\citep{multilingual_curse2023}, implies that an over-reliance on English training leads to sub-optimal performance for non-English languages, as the model lacks sufficient exposure to other languages during pre-training.

In this paper, we aim to develop LLMs that perform well across the South-East Asia (SEA) region, encompassing a range of languages that include English, Chinese, Vietnamese, Thai, Indonesian, Malay, and Lao.
We share both successful experiences and failed attempts in a completely open manner to accelerate the development of LLMs for the SEA region.
Specifically, we introduce and discuss the benefits of merging adjacent short example, the document-level code-switching and the word-level code-switching, as illustrated in Figure~\ref{fig:sailor_overview}.
Additionally, we share all of our data cleaning pipeline and deduplication procedure that turns out to be extremely important for the quality of LLMs, especially in the scenario of continual pre-training.
As for the tokenization, we explore the usage of BPE Dropout~\citep{provilkov-etal-2020-bpe} and highlight its importance for the robustness of LLMs.
Finally, we use small models as proxy to optimize the hyper-parameter for continual pre-training, including the learning rate and data mixture ratio of different data sources.

\section{Insights}

\begin{table}[b]
\small
    \centering
    \begin{tabular}{cccc}
    \toprule
    \textbf{Technique} & \textbf{Stage} & \textbf{Used} & \textbf{Note} \\
    \midrule
    Merging Adjacent Short Examples & Data & Yes & Improve Performance \\
    Document-Level Code-Switching & Data & Yes & Improve Performance \\
    Word-Level Code-Switching & Data & No & Marginal Effect w. Document-Level \\
    Aggressive Data Deduplication & Data & Yes & Improve Performance \\
    Aggressive Data Cleaning & Data & Yes & Improve Performance \\
    Vocabulary Expansion & Tokenization & No & Challenging to Apply \\
    BPE Dropout & Tokenization & Yes & Improve Robustness \\
    Learning Rate Tuning & Training & Yes & Accelerate the Training \\
    Data Mixture Simulation & Training & Yes & Balance Different Languages \\
    \bottomrule
    \end{tabular}
    \caption{The techniques we mainly consider during our development.}
    \label{tab:highligh_tech}
\end{table}

During our development, we perform ablation studies on small LMs to understand the impact of various strategies. We then apply the key insights gained from these studies to improve LLMs.
Most of the experimental results are obtained from three series of models: our internal 120M model trained on 20B English tokens using SlimPajama~\citep{cerebras2023slimpajama}, the TinyLlama 1.1B model~\citep{tinyllama2024}, and the Qwen1.5-0.5B model~\citep{qwen}.
All techniques we have considered are listed in Table~\ref{tab:highligh_tech}.

\subsection{Data}

\paragraph{Merging Adjacent Short Examples} 
Several studies have emphasized the importance of deduplication~\citep{lee-etal-2022-deduplicating}, with some popular corpora undergoing deduplication at the paragraph level, resulting in a final corpus comprising unique paragraphs such as CC100~\citep{CC1002020}.
The approach enhances data efficiency by maximizing the number of unique tokens the model encounters, but it can adversely impact model performance as the connections between different pieces within the context become less relevant.
To mitigate the issue, we have employed a simple method of randomly combining several adjacent examples into one example before applying a global shuffle.
The method can be applied because the deduplicated paragraphs still retain the order in which they appear in the original documents, allowing for the reconstruction of context when necessary.
Moreover, the method is also applied to certain sources, such as subtitles, which are inherently composed of short sentences.

\paragraph{Code-Switching} Code-switching refers to the phenomenon where different languages are used within the same context. We categorize code-switching into two types: \textit{document-level} and \textit{word-level}. For document-level code-switching, when preparing the dataset for pre-training sequences, we pack documents from various languages instead of separately packing them within each language. Regarding word-level code-switching, we randomly select some words in each documents written in SEA languages (e.g., Indonesian) and replace 10\% of them with their corresponding English phrases, if available.
Our preliminary experiments on TinyLlama show that the document-level code-switching alone performs better than the word-level code-switching alone or a combination of both approaches.
Therefore, we only apply the document-level code-switching method during continual pre-training.
Interestingly, despite the intuitive expectation that incorporating translation data, such as CCAligned~\citep{elkishky_ccaligned_2020}, would enhance model performance on document-level code-switching, the experimental results did not demonstrate a significant improvement.
In contrast, using translation data alone (i.e., without document-level code-switching) can lead to improved model performance over the baseline on general tasks (e.g., question answering), suggesting that translation data plays a role similar to document-level code-switching to a certain degree.
Nonetheless, to facilitate translation capabilities, our dataset incorporates translation datasets, which will be discussed later.

\paragraph{Aggressive Data Cleaning and Deduplication}
The data quality is crucial during continual pre-training. 
We employ aggressive cleaning parameters, extended filtering list for each language, and multi-round deduplication.
Consequently, even though we started with well-curated open datasets, e.g., MADLAD-400 clean set~\citep{MADLAD2023}, we still further removed $31.11$\% in data cleaning and  $11.16$\% in data deduplication.
By extensively filtering out noisy, harmful, and duplicated content, we are able to significantly improve the efficiency of the pre-training process and the stability of the optimization procedure.
Furthermore, LLMs are less prone to memorization issues when training data has undergone thorough deduplication~\citep{lee-etal-2022-deduplicating}.

\subsection{Tokenization}

\begin{figure*}[b]
    \centering
    \begin{subfigure}[b]{0.45\textwidth}
        \includegraphics[width=\textwidth]{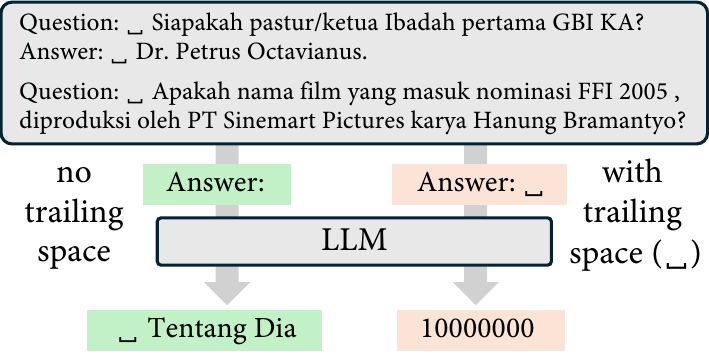}
        \caption{Minor variations in prompts such as a trailing space visualized by ␣ can drastically change the prediction of LLMs.}
        \label{fig:bpedropout_fig_a}
    \end{subfigure}
    \hfill
    \begin{subfigure}{0.51\textwidth}
        \centering
        \small
        \begin{tabular}{lcc}
            \toprule
            \textbf{Ablation} & \textbf{Prompt} & \textbf{Exact Match} \\
            \midrule
            \multirow{2}{*}{Sailor-1.8B} & no space & 40.88 \\
             & with space & 38.41 \\
            \midrule
            \multirow{2}{*}{\textit{w.o.} BPE dropout} & no space & 38.94 \\
            & with space & 18.76 \\
            \bottomrule
        \end{tabular}
        \caption{Experimental results on the TydiQA dataset indicate that applying BPE dropout significantly enhances the robustness of the Sailor-1.8B model when handling trailing spaces.}
        \label{fig:bpedropout_fig_b}
    \end{subfigure}
    \caption{
    Initially Sailor models were trained on 200B tokens using a greedy tokenization strategy.
    Subsequently, they were fine-tuned using BPE dropout for an additional 2B tokens, with the dropout rate as 0.1. As observed, BPE dropout improves the robustness.
    }
    \label{fig:bpedropout_tydiqa}
\end{figure*}

\paragraph{BPE Dropout} We have observed that the model is unreasonably sensitive to small variations of the prompt, especially on \textit{spaces}. As illustrated in Figure~\ref{fig:bpedropout_fig_a},
when prompting the model with the string ``Answer:'' without any trailing space yields a substantially improved performance compared to ``Answer: ␣''~\footnote{We use ``␣'' to represent space.}.
The same phenomenon are observed in Qwen1.5, Mistral and Llama 2, and a similar issue has been discussed at lm-evaluation-harness library\footnote{\url{https://github.com/EleutherAI/lm-evaluation-harness/issues/614}}~\citep{eval-harness}.
We attribute this kind of vulnerability to the tokenization strategy used in data processing.
Modern tokenization methods usually employ the Byte Pair Encoding (BPE)~\citep{sennrich-etal-2016-neural} under the greedy segmentation setting~\footnote{The default BPE class is initialized with no dropout in the HuggingFace \texttt{tokenizers} library.}, which means that sentences are segmented into subwords using the optimal tokenization strategy. 
However, the always-optimal strategy can lead to vulnerability in the model when it encounters noisy subwords, such as an unexpected space in ``Answer: ␣''.
Typically, a space is segmented into subwords alongside the subsequent chars (e.g., ``␣1'' constitutes a single subword).
Yet, if a space is left at the end of the prompt, it becomes an isolated subword ``␣'', deviating from the segmentation strategy in the demonstration examples.
To alleviate the problem, we employ the BPE-Dropout~\citep{provilkov-etal-2020-bpe} during continual pre-training, which stochastically corrupts the segmentation procedure of BPE to achieve subword regularization.
Experimental results indicate that although BPE Dropout slightly increases the loss on greedy subword segmentation, it enhances both the performance and the robustness of models, as shown in Figure~\ref{fig:bpedropout_fig_b}.

\begin{figure}[t]
    \centering
    \includegraphics[width=0.5\textwidth]{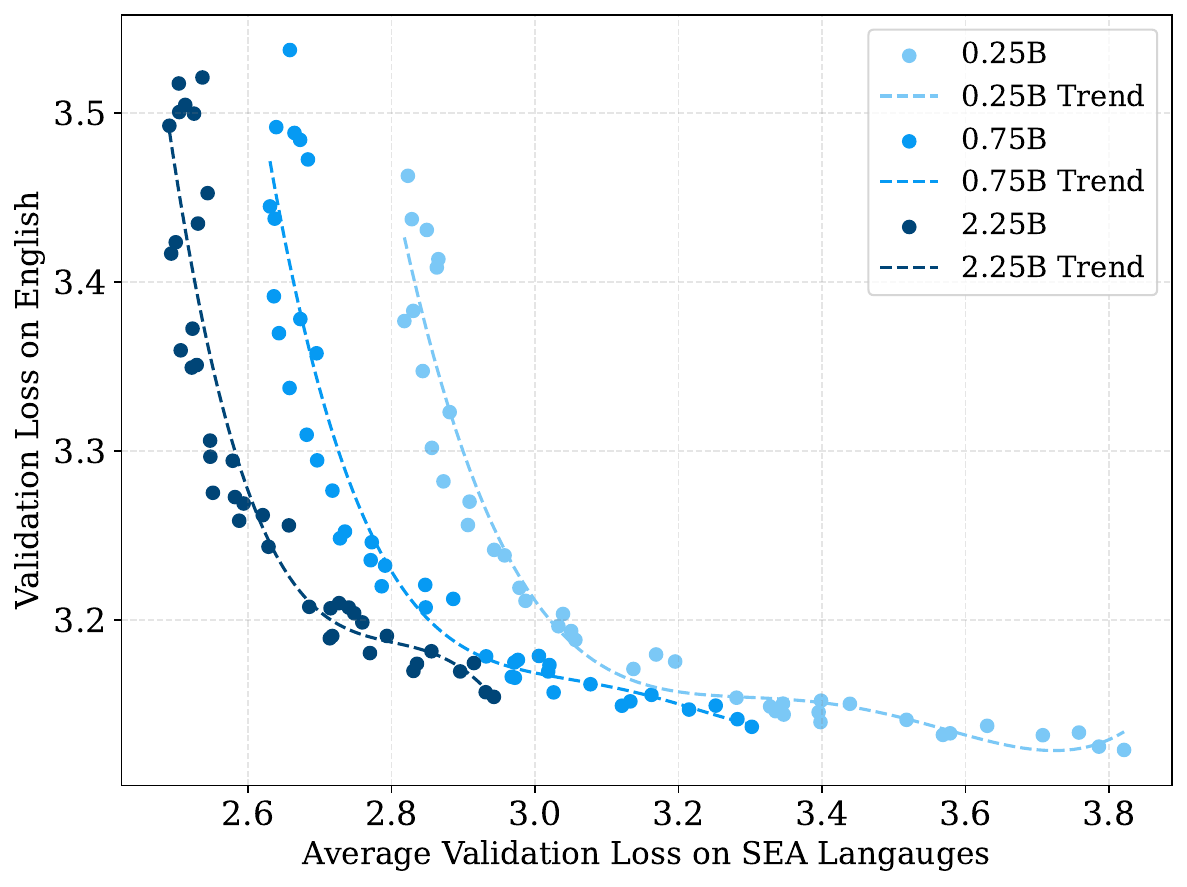}
    \caption{We initially pre-train a 120M model using a corpus of 20B tokens focusing on English. Subsequently, we continually pre-train the model using a mixed corpus comprising both English and SEA languages. Each data point here corresponds to a different configuration of data mixture and learning rate. As indicated, under a fixed total tokens, there is a trade-off between the model's performance on English and SEA languages.}
    \label{fig:en_sea_loss}
\end{figure}

\paragraph{Vocabulary Expansion} 
We have tried our best to do vocabulary expansion on models like Mistral~\citep{mistralpaper} and Llama-2~\citep{llama2paper}. 
However, similar to the observation in concurrent works~\citep{llamabeyond2024}, it is challenging to expand the vocabulary with maintaining the original performance.
According to our investigation, without sufficient continual pre-training, the performance of vocab expanded model could not even recover to the baseline version.
For example, after being trained on 20B tokens with an expanded vocabulary of 15,000 subwords, Mistral's question answering performance over Thai remains 10\% lower than that of the original model.
We have also explored several methods to eliminate the problem, including warmup the embedding layer first, or modularized continual-training~\citep{Kim2024EfficientAE}.
Despite our efforts, the methods did not perform as effectively as we expected.
We acknowledge this interesting yet challenging problem as an opportunity for future research.
Finally, we decided to develop Sailor models based on Qwen1.5~\citep{qwen}, which is inherently multilingual-friendly and possesses a large vocabulary size, thereby guaranteeing a high compression rate for SEA languages.

\subsection{Training}\label{insight:training}

When it comes to continual pre-training, two crucial hyper-parameters to consider are the learning rate and the data mixture.
In our practise, we begin by generating a number of training configurations with varying learning rates~\footnote{We first divide the logarithmic range between 1e-5 and 4e-4 into 20 equal intervals. For each configuration, we randomly select one interval and use the corresponding value as the learning rate.} and language proportions to train several proxy models~\footnote{The proxy models are typically small, allowing for cheap experiments, yet they retain key characteristics akin to those of the target base LLM.}.
By analyzing the trade-off between English and SEA languages on these proxy models, we can select a suitable learning rate.
Once the learning rate is determined, we then conduct fine-grained data mixture simulation experiments to optimize the joint loss across all languages, which is finally used in large-scale training.

\begin{figure}[t]
    \centering
    \begin{subfigure}[b]{0.47\textwidth}
        \includegraphics[width=\textwidth]{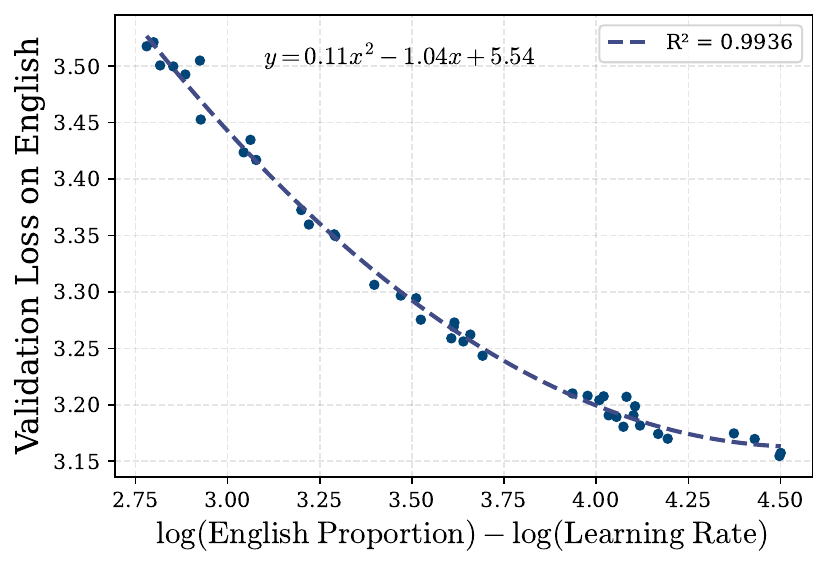}
        \caption{The relationship between English loss and $\log(\textrm{English Proportion}) - \log(\textrm{Learning Rate})$.}
        \label{fig:eng_malay_lr_a}
    \end{subfigure}
    \hfill
    \begin{subfigure}[b]{0.47\textwidth}
        \includegraphics[width=\textwidth]{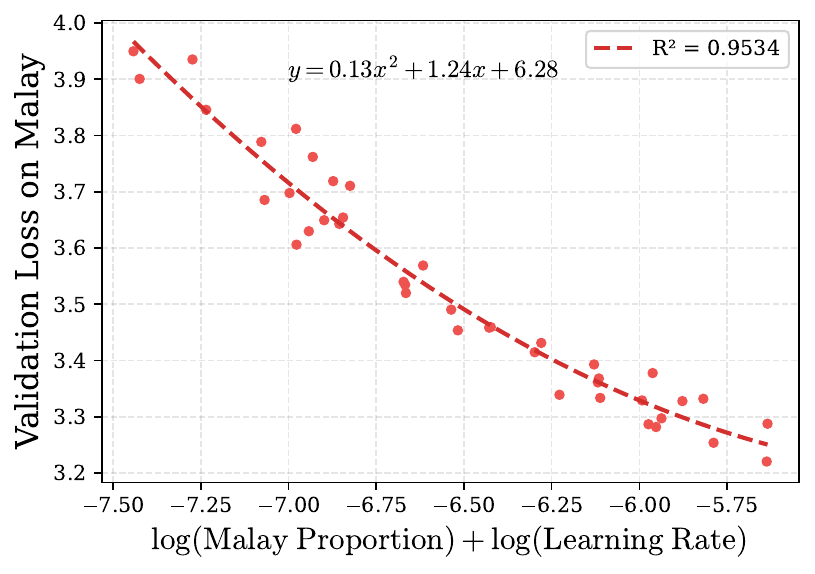}
        \caption{The relationship between Malay loss and $\log(\textrm{Malay Proportion}) + \log(\textrm{Learning Rate})$.}
        \label{fig:eng_malay_lr_b}
    \end{subfigure}
    \begin{subfigure}[b]{0.57\textwidth}
        \includegraphics[width=\textwidth]{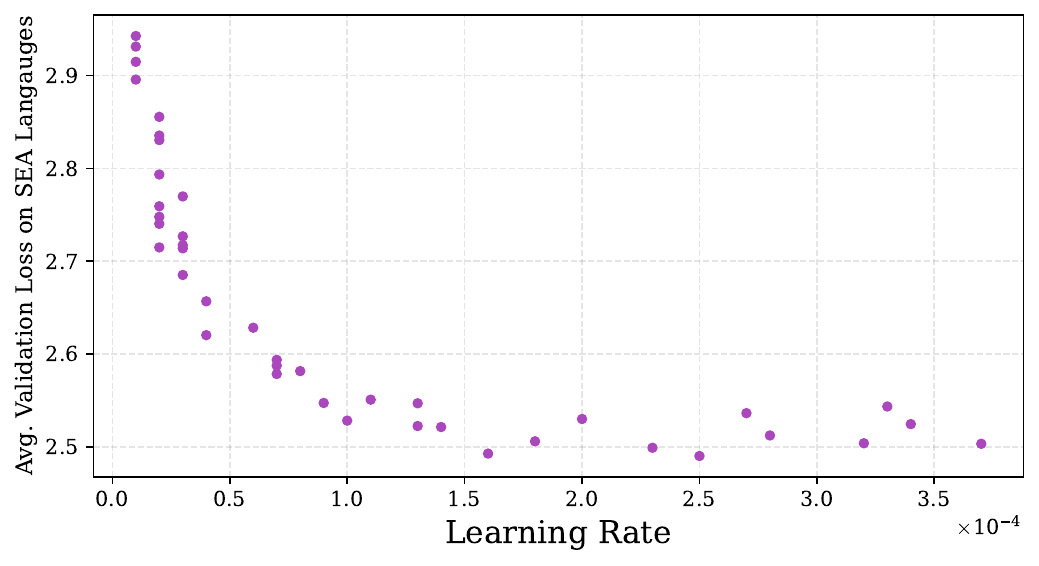}
        \caption{The average SEA loss with increasing the learning rate.}
        \label{fig:eng_malay_lr_c}
    \end{subfigure}
    \caption{Under the same token budget, we observe that (a) the validation loss on English can be modeled as a quadratic function of $\log(\textrm{English Proportion})-\log(\textrm{Learning Rate})$; (b) the validation loss on SEA languages, using Malay as an example, can be approximately represented by a quadratic function with $\log(\textrm{Malay Proportion})+\log(\textrm{Learning Rate})$;
    (c) we can tune the learning rate by analyzing the learning curves on SEA languages.}
    \label{fig:eng_malay_lr}
\end{figure}

\paragraph{The Curse of Multilinguality}

Figure~\ref{fig:en_sea_loss} provides a visual representation of the relationship between the model's performance on English and SEA languages when subjected to the same token budget (e.g., 0.25B).
It clearly illustrates that the trade-off that exists among different languages.
In other words, when performing continual pre-training on an English-centric language model, increasing the proportion of SEA language corpus always results in a degradation of the model's performance on English, even when a high-quality English corpus is included for replay.
These findings align with previous studies on the curse of multilinguality~\citep{conneau-etal-2020-unsupervised,multilingual_curse2023}, which posit that modeling multiple languages within a single model leads to competition among languages for the fixed model capacity.

\paragraph{Learning Rate Tuning}

Figure~\ref{fig:en_sea_loss} also demonstrates an inverse relationship between the number of tokens and the loss on English.
As more tokens are consumed (e.g., 0.25B $\rightarrow$ 2.25B), the curve shifts towards the upper-left area, signifying an increase in the loss on English.
Interestingly, the loss trend on the source domain (i.e., English) is primarily influenced by two factors: the proportion of English data during continual pre-training and the learning rate.
Under the same token budget, the model's loss on English can be accurately modeled as a quadratic function of $\log(\textrm{English Proportion})-\log(\textrm{Learning Rate})$, as shown in Figure~\ref{fig:eng_malay_lr_a}.
In other words, while keeping the proportion of English data constant, increasing the learning rate may adversely affect the model's performance on English.

Meanwhile, the loss trend on the target domain (i.e., SEA languages) is also mainly affected by the proportion of the target domain and the learning rate.
However, there is a different modeling among the model loss on SEA languages, the proportion and the learning rate, as demonstrated by Figure~\ref{fig:eng_malay_lr_b}.
From the observation, it becomes evident that the learning rate serves as a crucial hyper-parameter.
A well-tuned learning rate plays a pivotal role in striking a balance between the acquisition of SEA languages and the forgetting of English.
As shown in Figure~\ref{fig:eng_malay_lr_c}, considering that increasing the learning rate beyond 1e-4 does not yield significant improvements in the loss on SEA languages, we set the peak learning rate to 1e-4 in our experiments.

\paragraph{Data Mixture Simulation} 

\begin{figure}[t]
    \centering
    \includegraphics[width=1.0\columnwidth]{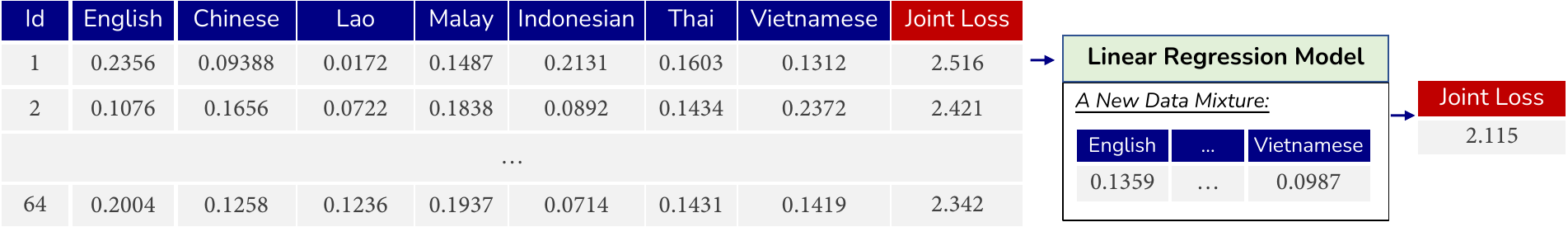}
    \caption{We employ the experimental results from proxy models across a variety of data mixtures (e.g., 64 distinct data mixture here) to fit a linear regression model. The model is then utilized to predict the validation loss of simulate numerous random data mixtures, enabling us to identify the most effective data mixture for optimizing joint loss. Subsequently, the best data mixture is applied to large-scale training.}
    \label{fig:linear_regression_model}
\end{figure}

We aim to develop an improved LLM tailored for the entire SEA region, with a focus on ensuring balanced representation across all target languages. To achieve this, we have developed a new algorithm that determines the appropriate weights for various languages during continual pre-training. This method involves conducting a series of randomized data mixture experiments, while adhering to a predetermined learning rate.
Our goal is to determine the most effective data mixture.
To this end, we suggest employing simulations in conjunction with linear regression models.
As depicted in Figure~\ref{fig:linear_regression_model}, we begin by training a set of proxy models (e.g., 64 in total here) on a variety of data mixtures for a limited number of training steps (e.g., 1000 steps).
We then fit a linear regression model, using the data mixture as the input feature and the joint loss considering all languages~\footnote{We use the product of individual losses as the joint loss.} as the target.
With this model, we can perform numerous simulation experiments (e.g., 1,000,000) on randomly sampled data mixtures to explore the vast array of possibilities within seconds.
The linear model then guides us in selecting the combination that yields the lowest predicted joint loss.
Once this data mixture has been optimized, it can be directly applied to large-scale training.
More details and findings will be discussed in our upcoming paper.

\subsection{Best Practise for Continual Pre-training}

Drawing from the above insights, we highlight the importance of selecting the learning rate and the proportion of source domain data to mitigate issues such as catastrophic forgetting.
Therefore, we focus on the metric $\log(\textrm{Source Domain Proportion}) - \log(\textrm{Learning Rate})$, which we refer to as the \textit{magic metric} below. We suggest the following steps:
\begin{enumerate}
    \item Fit a parametric quadratic function modeling the relationship between loss$_{~\textrm{source}}$ and the magic metric via experiments varying learning rates and proportions.
    \item Estimate the boundary of the magic metric value beyond which the model's loss$_{~\textrm{source}}$ starts to deviate significantly from the original one.
    \item Balance the learning progress on the target domain with the retention rate on the source domain by selecting a suitable magic metric larger than the boundary.
    \item If the magic metric substantially exceeds the estimated boundary, it indicates that the model retains more knowledge from the source domain; conversely, it facilitates a more rapid learning pace on the target domain.
\end{enumerate}

The above guideline can potentially explain why Lemur~\citep{xu2024lemur} demonstrated negligible performance deterioration on natural language benchmarks (e.g., MMLU) despite undergoing continual pre-training from Llama-2 on an extremely imbalanced data distribution (i.e., text:code as 1:10). The employment of a smaller learning rate (i.e., 4e-5) during Lemur's training likely preserved the magic metric within an good range, allowing the model to maintain its proficiency in the source natural language domain.

\section{Data Sources}

Here we describe all the corpus used in our training.
Note that we performed an additional round of data deduplication and cleaning on these datasets before using them.

\subsection{Dataset for Replay}

To mitigate catastrophic forgetting of English and Chinese capabilities of our models, we consider high-quality English and Chinese datasets as part of our data sources during continual pre-training.

\paragraph{SlimPajama}
SlimPajama~\citep{cerebras2023slimpajama} is a high-quality dataset comprising 627B tokens, curated by rigorously cleaning and deduplicating the RedPajama Corpus~\citep{together2023redpajama}. It primarily focuses on English, and removes 49.6\% of low-quality and duplicate data from RedPajama.
We use the released version on HuggingFace~\footnote{\url{https://huggingface.co/datasets/cerebras/SlimPajama-627B}}.

\paragraph{SkyPile}

SkyPile~\citep{skywork2023} is a massive, high-quality Chinese dataset for pre-training. It comprises 233M web pages, totaling 150B tokens, carefully filtered and deduplicated from public web sources. We download SkyPile by accessing its hosted dataset on HuggingFace~\footnote{\url{https://huggingface.co/datasets/Skywork/SkyPile-150B}}.

\subsection{Dataset for SEA Languages}

\paragraph{CC100} CC100~\footnote{\url{https://data.statmt.org/cc-100}} is a multilingual corpus comprising monolingual data from over $100$ languages. The corpus was original constructed for training the XLM-R model~\citep{conneau-etal-2020-unsupervised}, a powerful cross-lingual language model. The data was sourced from the Common Crawl project~\citep{CommonCrawl2018}. Specifically, the corpus was generated by processing Common Crawl snapshots from January to December 2018, using the open-source CC-Net repository~\citep{CC1002020}. In our pre-training corpus, we take the Indonesian, Malay, Lao, Thai and Vietnamese subsets.

\paragraph{MADLAD-400} The CC100 corpus is a great resource for multilingual languages due to its high quality, but it has already split every document into separate paragraphs, making it serve as a paragraph-level corpus. We believe using paragraphs as examples would greatly hurt the document-level performance of the model, as evidenced by our preliminary study. Therefore, we also consider MADLAD-400~\citep{MADLAD2023}, a manually audited and large-scale multilingual corpus spanning 419 languages. MADLAD-400 is also based on CommonCrawl, which uses all available corpus till August 2022. In our pre-training corpus, we take its clean version, downloaded from the dataset hosted by HuggingFace~\footnote{\url{https://huggingface.co/datasets/allenai/MADLAD-400}}.

\paragraph{Wikipedia} We utilize the Wikipedia dump (encompassing Malay, Indonesian, Thai, and Vietnamese) up to November 2023 from the Wikipedia dataset hosted on HuggingFace~\footnote{\url{https://huggingface.co/datasets/wikimedia/wikipedia}}. It should be noted that some of the Wikipedia corpus may be duplicated, as the SlimPajama dataset has already included the multilingual Wikipedia corpora.

\paragraph{OpenSubtitles}

We collect the Malay, Indonesian, Thai and Vietnamese subtitles from the OPUS OpenSubtitles category~\footnote{\url{https://opus.nlpl.eu/OpenSubtitles-v2018.php}}. For all subtitles, we use a sliding window of $100$ to concatenate adjacent subtitles to compose longer documents. An example of Indonesian subtitle can be found below:

\begin{tcolorbox}
Duduk Manis dan Selamat Menikmati \\
Su-ho!\\
Asapnya terus mendekatiku. \\
Kembali ke kampung halaman, Kukira aku akan dapat Sashimi...\\
$\cdots$
\end{tcolorbox}

\paragraph{Translation} While our preliminary studies indicate that translation data may have similar effects to document-level code-switching, we still incorporated translation data since translation is an important task.
We curate a selection of English-SEA language translation pairs available in the OPUS project~\footnote{\url{https://opus.nlpl.eu/}} (e.g., TED2020 talks). Notably, we observe substantial duplication within the translation data, thus necessitating a further deduplication step.
Concurrently, to account for both directions, we processed data for both English-to-SEA and SEA-to-English translation directions for each example. An illustrative example is provided below:

\begin{tcolorbox}
Indonesian to English: Pak Tanaka bukan murid. Mr. Tanaka is not a student.\\
English to Indonesian: Did the Israelites execute criminals by hanging them on stakes? Apakah mereka menghukum mati penjahat dengan memakukannya pada tiang?
\end{tcolorbox}

\section{Preprocessing Pipeline}

\begin{figure}[t]
\centering
\includegraphics[width=0.8\textwidth]{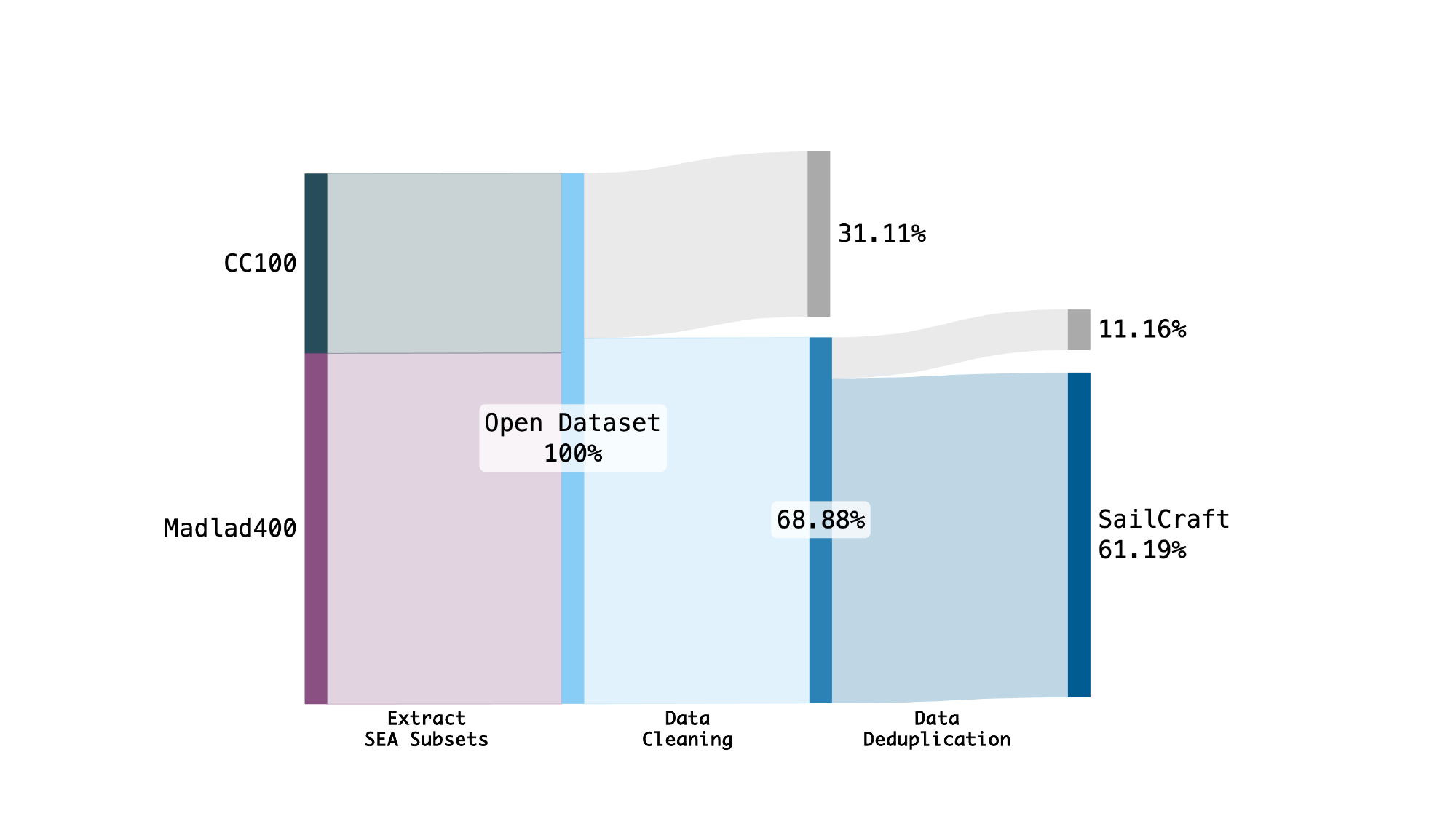}
\caption{
With aggressive data cleaning and deduplication, we obtain 61.19\% high-quality data from two well-curated open datasets, including CC100~\citep{CC1002020} and MADLAD-400~\citep{MADLAD2023}.
This forms the \textbf{SailCraft} dataset, used to train the \textbf{Sailor} models.
The reported removal rate (grey) is with respect to each previous stage, and the kept rate (colored) demonstrates the overall rate.
}
\label{fig:data_clean_rate}
\end{figure}

The data quality is crucial during continual pre-training. 
We found that several publicly available multilingual datasets could be further cleaned and deduplicated.
To improve the data cleaning process for SEA languages specifically, we expanded our list of filtering words, trained new filtering models, and implemented a more aggressive deduplication strategy. 
As a result of these optimizations, we extracted 61.19\% of data for SEA languages from public datasets, and constructed the final \textit{SailCraft} dataset.
The specific removal rates are shown in Figure~\ref{fig:data_clean_rate}.

\subsection{Data Normalization}

We apply the following data normalization procedures before data cleaning:
\begin{enumerate}
    \item \textbf{Uniform whitespace.} 
    We first unify the whitespace within the sentence, transforming all forms of whitespace to the classic space character. 
    This approach guarantees consistency across various whitespace characters and facilitates the segmentation, i.e., converting the documents into words.
    
    \item \textbf{Replace Unicode punctuation.} 
    We replace Unicode punctuation in text with ASCII equivalents. 
    It ensures the compatibility by simplifying text processing, as ASCII punctuation is more universally recognized and easier to work with.
    
    \item \textbf{Remove incorrect words.}
    We exclude emojis using the \textsc{emoji} package~\footnote{\url{https://github.com/carpedm20/emoji}}, remove HTML-related tags to eliminate links associated with source page code, and filter out certain terms based on a pre-defined word list.
    
    \item \textbf{Remove lengthy words.} 
    We remove words exceeding a pre-defined length cutoff, especially in web-scraped datasets where lengthy words often signify formatting errors or URLs. This process helps to clean and standardize the data.
    
\end{enumerate}

Note that for MADLAD-400, we have fixed the Unicode escaping issue (i.e., lots of ``\textbackslash \textbackslash n'') as introduced in HuggingFace forum~\footnote{\url{https://huggingface.co/datasets/allenai/MADLAD-400/discussions/2}}, which would cause the trouble in few-shot in-context learning and chatbot application where ``\textbackslash n'' acts as an important delimiter between demonstrations and task input.
Concretely, we replace all ``\textbackslash \textbackslash n'' with ``\textbackslash n\textbackslash n'' or ``\textbackslash n'' with some heuristic rules. Please refer to Appendix~\ref{appendix:fix_madlad} for implementation details and concrete cases.

\subsection{Data Cleaning}

The data cleaning mainly follows the BigScience data cleaning recipe~\footnote{\url{https://drive.google.com/file/d/1cCJ8sWE88TRLDAa3eHLmXO4JlkR2QzLY/view}}.
Note that for most languages, we can make use of the publicly available resource. 
However, for several low-resource languages, we have to train model from scratch.
The entire data cleaning process is as follows:

\begin{enumerate}
    \item \textbf{Filtering on the number of words.}
    We first tokenize the documents with sentencepiece model for each language.
    Then we count the number of words and remove the document that is less than \textbf{min\_length} or is greater than the \textbf{max\_length}.
    The purpose of filtering short document is to remove the incorrect sentences or the sentences without enough context.
    The purpose of filtering long document is to remove redundant information or exceed the maximum input length.
    
    \item \textbf{Filtering on the character repetition ratio.}    
    We first compile the list of character-level $n$-grams for the given document. 
    Then, we calculate the frequency for each $n$-gram. 
    We define the character repetition ratio as the sum of frequencies of the top $m$ most frequent $n$-grams. 
    A document is dropped if its character repetition ratio score is above the pre-defined threshold.
    Note that $m$ is determined as a trade-off choice so that it can balance the distribution of short and long documents. Practically, we choose $m$ as the square root of the amount of $n$-grams.
        
    \item \textbf{Filtering on the word repetition ratio.}
    The word repetition ratio is defined as the sum of frequencies of all $n$-grams whose frequency is greater than 2.
    A document is dropped if its word repetition ratio score is above the pre-defined threshold.
    
    \item \textbf{Filtering on the special characters ratio}.
    A list is maintained to track special characters.
    If a document's ratio of special characters exceeds a pre-defined threshold, it will be dropped. The purpose of this filter is to eliminate documents that consist primarily of special characters.
    
    \item \textbf{Filtering on the stop words ratio.}
    A list of stop words for each language is maintained, and a document will be removed if its stop words ratio is above the pre-defined threshold.
    It is to remove machine-generated text that do not have much semantically meaningful information.
    However, one significant challenge arises with languages such as Chinese and Vietnamese that do not use spaces, as it becomes difficult to recognize stop words after tokenization. 
    Following BigScience practise, we address the issue by expanding the stop list to include both word-level and byte-piece-level stop words, thereby enhancing the coverage and effectiveness of the filtering.
    For stop words list, we collected those for Thai~\footnote{\url{https://github.com/stopwords-iso/stopwords-th/blob/master/stopwords-th.txt}} and Malay~\footnote{\url{https://github.com/stopwords-iso/stopwords-ms/blob/master/stopwords-ms.txt}} from available resources.
    However, we did not find relevant resources for Lao, and thus we translated the Thai stop words list into Lao.
    
    \item \textbf{Filtering on the flagged words ratio.}
    We maintain a list of flagged words for each language.
    A document is removed if its flagged words ratio is above the pre-defined threshold.
    It will remove buzzwords about the porn, which is harmful for model training.
    We create or expand the flagged word list for Thai, Malay, and Lao by translating from English ones developed by BigScience.

    \item \textbf{Filtering on the language identification prediction score.}
    We adopt the fastText~\citep{joulin2016fasttext} model~\footnote{\url{https://dl.fbaipublicfiles.com/fasttext/supervised-models/lid.176.bin}} to get the language identification result for each document and the corresponding confidence score.
    A document will be dropped if its confidence score is below the pre-defined threshold.
    The filter is to remove unnatural content such as machine-generated text, advertisements, or frequently changing spoken language.
    However, it also bring the drawback that it would remove code-switching text that exists in SEA regions like Singlish (Singapore English) and Manglish (Malaysian English).
    
    \item \textbf{Filtering on the perplexity score.}
    We adopt the KenLM~\citep{heafield-2011-kenlm} model to calculate the perplexity score of documents for each language.  
    KenLM model are trained from the high-quality corpus like Wikipedia.
    A document will be removed if its perplexity score is above the pre-defined threshold. 
    The filter will remove the documents with unrelated words like tags, time, date and lots of repetitions.
    One main drawback of the filter is that it would inevitably remove necessary documents that have a different distribution from Wikipedia.
    For KenLM model, we download most language models from BigScience repo~\footnote{\url{https://github.com/bigscience-workshop/data-preparation/tree/main/preprocessing/training/01b_oscar_cleaning_and_filtering\#2-download-everything-you-need}}.
    However, there is no KenLM model available for Thai, Malay and Lao.
    Thus, we sample a high-quality subset from the Wikipedia corpus and train KenLM models with vocab size 65536~\footnote{\url{https://github.com/bigscience-workshop/data_tooling/tree/master/kenlm_training}}.
    
\end{enumerate}

\subsection{Data Deduplication}

\begin{table}[t]
\centering
\begin{tabular}{llrrr}
\toprule
\textbf{Language} & \textbf{Source} & \textbf{Raw} & \textbf{After Clean} & \textbf{After Dedup} \\
\midrule
\multirow{2}{*}{Indonesian} & CC100 & 149G & 105G & 88G \\
 & MADLAD-400 & 140G & 130G & 126G \\
\midrule
\multirow{2}{*}{Thai} & CC100 & 72G & 33G & 11G \\
 & MADLAD-400 & 283G & 103G & 94G \\
\midrule
\multirow{2}{*}{Vietnamese} & CC100 & 138G & 95G & 77G \\
 & MADLAD-400 & 281G & 262G & 251G \\
\midrule
\multirow{2}{*}{Malay} & CC100 & 8.5G & 7.2G & 4.8G \\
 & MADLAD-400 & 12.0G & 12.0G & 12.0G \\
\midrule
\multirow{2}{*}{Lao} & CC100 & 0.62G & 0.15G & 0.06G \\
 & MADLAD-400 & 1.80G & 0.74G & 0.69G \\
\bottomrule
\end{tabular}
\caption{The storage statistics on each subset, including raw data, after data cleaning and after data deduplication. Even though we started with high-quality open datasets (i.e., the MADLAD-400 clean set and the CC100), we still removed 31.11\% data during data cleaning, and further removed 11.16\% during data deduplication.}
\label{tab:data_clean_open}
\end{table}

The data deduplication procedure is the most important and challenging part in our data preprocessing.
Firstly, it distills the corpus for efficient pre-training.
Moreover, it further filters the noise information for effective training, like machine-generated advertisements that could not be easily recognized by rule-based cleaning methods.
Most importantly, LLMs are less prone to exhibit memorization issues when training data has undergone thorough deduplication~\citep{lee-etal-2022-deduplicating}.

\begin{figure}[tb]
\centering
\includegraphics[width=1\textwidth]{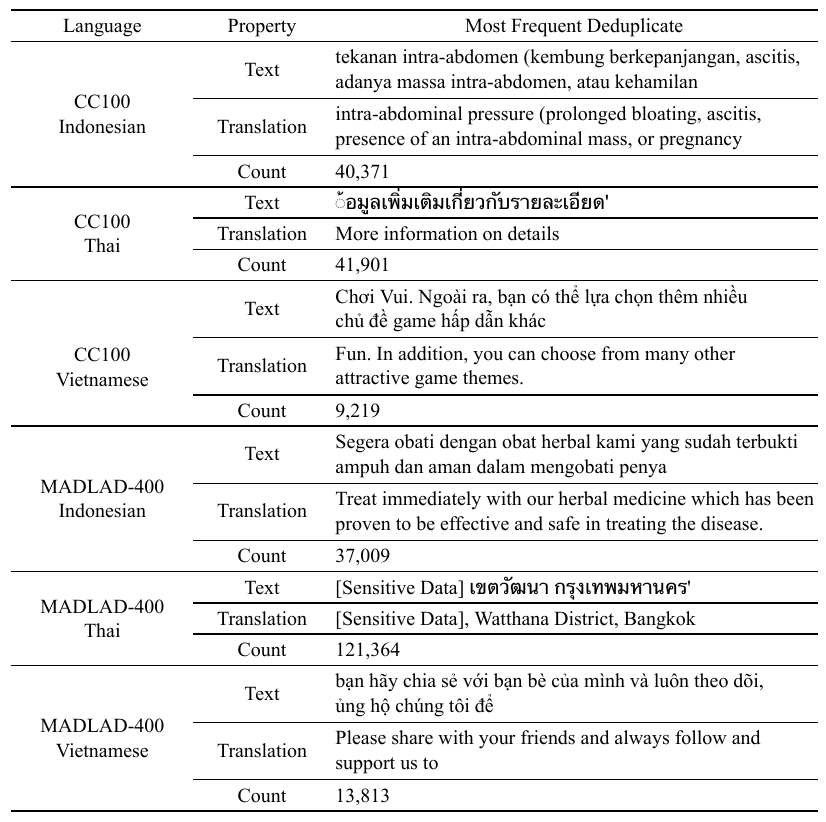}
\caption{
The most frequent textual duplicates identified across two datasets for three SEA languages, along with their respective frequencies.
Inappropriate content and personally identifiable information are replaced with [Sensitive Data].
For brevity, we highlight only the duplicate content.
The lengthy content is truncated for better visualization.
}
\label{fig:dedup-top1}
\end{figure}

\begin{figure}[tb]
\centering
\includegraphics[width=1\textwidth]{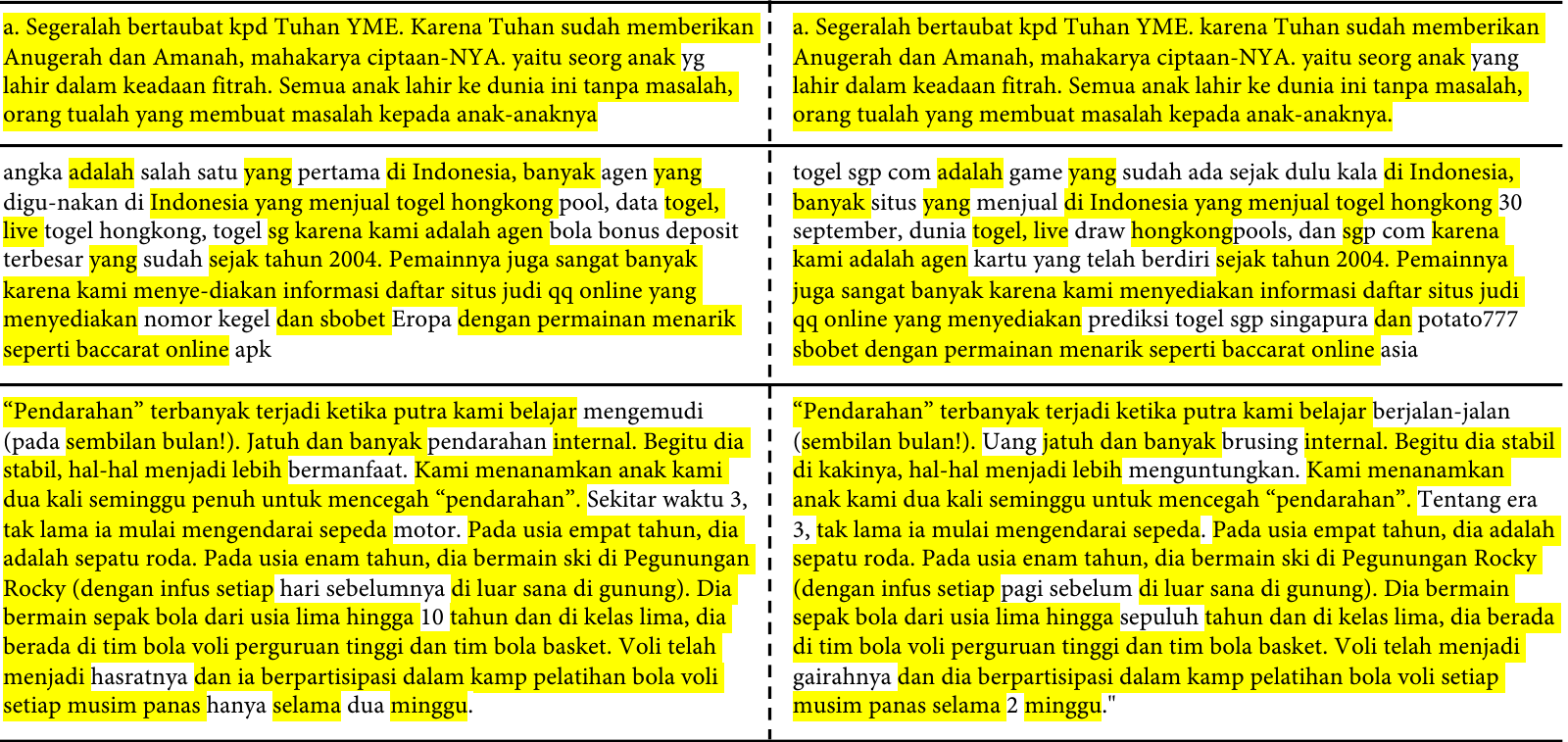}
\caption{
The above pairs of documents from the CC100-Indonesian dataset are identified as duplicates by our deduplication algorithm. 
To enhance readability, the matching sub-sequences within these document pairs are highlighted.
}
\label{fig:data-deduplication-example}
\end{figure}

\paragraph{Implementation Details}
For implementation, we employ the Text-Dedup tool~\citep{chenghao_mou_2023_8364980} for data deduplication~\footnote{\url{https://github.com/ChenghaoMou/text-dedup}}. It is well-packaged and offers easy command-line interaction. The tool also demonstrates excellent performance in deduplication benchmarking~\footnote{\url{https://tinyurl.com/bdf6zerm}}. It utilizes 5-gram MinHashLSH deduplication~\citep{Broder1997OnTR} with a threshold of 0.7.
The number of permutations (hashes) is set to 256 to conserve memory usage. The Jaccard similarity threshold is set to 0.7. 
The Text\-Dedup tool further computes the optimal MinHashLSH parameter that minimizes the weighted sum of probabilities of false positives and false negatives~\footnote{Refer to the \textsc{optimal\_param} function at \url{https://github.com/ChenghaoMou/text-dedup/blob/main/text\_dedup/utils/analysis.py}}. 
Ultimately, the number of bands and the number of rows per band (i.e., two crucial hyper-parameters) are optimized to 25 and 10, respectively.

For resource requirements, the data deduplication mainly consumes memory, CPU cores and disk spaces. 
The memory requirement is primarily determined by the number of documents, rather than the disk size of the documents~\footnote{\url{https://huggingface.co/blog/dedup}}. 
To facilitate dealing with large files within limited memory resource, we first split them into chunks (30GB) to make it tractable on our CPU server.
It takes about 200GB of memory to process a 30GB corpus with 256 permutations.
It takes approximately 30 minutes in total to deduplicate the 30GB corpus under 64 CPU cores.
To improve the deduplication performance, we iteratively cycled through the process of \textit{splitting into chunks, data deduplication, and recombining the chunks} for 3 rounds until the chunk size converged.

Definitely, more aggressive deduplication hyper-parameters (i.e., more permutations and larger chunk sizes) would further improve the accuracy of data deduplication. We aim to improve it from both algorithmic and engineering perspectives in the next version. 
For more discussion, please refer to the future work section.

\paragraph{Case Study}
Figure~\ref{fig:dedup-top1} showcases the most prevalent duplicate sentences across various language subsets identified by our deduplication algorithm. 
These duplicates span a wide range of domains, including medicine, customer service, and address information. 
The presence of such noisy and redundant data can impede the pre-training process, as indicated by \cite{refinedweb}. 
Additionally, sensitive information like emails and phone numbers poses privacy risks.

Our deduplication approach effectively addresses the prevalent scenario where documents are nearly identical, differing only in the interspersed template fields, as exemplified in Figure~\ref{fig:data-deduplication-example}. Despite cleaning efforts by CCNet~\citep{wenzek2020ccnet} and MADLAD-400~\citep{MADLAD2023}, the quality of open datasets remains sub-optimal, underscoring the challenges in multilingual data cleaning.
For more deduplication cases, please refer to Appendix~\ref{appendix:dedup}.

\subsection{Data Composition}\label{sec:data_composition}

\begin{table}[tb]
    \centering
    \small
    \begin{tabular}{cccc}
    \toprule
        \textbf{Language} & \textbf{Source} & \textbf{Effective Tokens (B)} & \textbf{Epoch} \\
        \midrule
        English & SlimPajama & 37.20 & 0.06 \\
        \midrule
        Chinese & SkyPile & 22.64 & 0.15 \\
        \midrule
        \multirow{2}{*}{Lao} & CC100 & 0.03 & 0.97 \\
        & MADLAD-400 & 0.31 & 0.97 \\
        \midrule
        \multirow{4}{*}{Malay} & CC100 & 2.02 & 1.34 \\
         & MADLAD-400 & 5.54 & 1.54 \\
         & OpenSubtitles & 0.04 & 1.07 \\
         & Wikipedia & 0.17 & 1.32 \\
        \midrule
        \multirow{5}{*}{Indonesian} & CC100 & 23.72 & 0.90 \\ 
         & MADLAD-400 & 25.62 & 0.66 \\
         & OpenSubtitles & 0.24 & 1.07 \\
         & Wikipedia & 0.45 & 1.32 \\
         & Translation & 0.50 & 1.16 \\
        \midrule
         \multirow{7}{*}{Thai} & CC100 & 3.00 & 1.28 \\
         & MADLAD-400 & 32.07 & 1.35 \\
         & OpenSubtitles & 0.13 & 1.01 \\
         & Wikipedia & 0.28 & 1.32 \\
         & Translation & 0.34 & 1.14 \\
        \midrule
        \multirow{5}{*}{Vietnamese} & CC100 & 14.25 & 0.82 \\
         & MADLAD-400 & 26.16 & 0.44 \\
         & OpenSubtitles & 0.05 & 1.08 \\
         & Wikipedia & 0.50 & 1.32 \\
         & Translation & 0.43 & 1.20 \\
        \bottomrule
    \end{tabular}
    \caption{The data composition of the final corpus SailCraft.}
    \label{tab:data_composition}
\end{table}

As detailed in Section~\ref{insight:training}, our algorithm involves utilizing proxy models to fit a linear regression model, which then aids in determining the optimal data mixture for large-scale training.
To elaborate, we extend the data mixture by extending it beyond language-level considerations to also include the source of the data.
This means we treat each language from every source as a distinct dataset and try to optimize the data mixture of these datasets.
The Qwen1.5-0.5B model serves as our proxy model, and we apply the optimized data mixture to the continual pre-training process across all model sizes.
The effective tokens and equivalent epochs in SailCraft are documented in Table~\ref{tab:data_composition}.
From the table, we observe that, in terms of quality or diversity, the CC100 dataset exhibits a relative advantage over the MADLAD-400 dataset, particularly for Indonesian and Vietnamese.

\section{Model Training}

We obtain the Sailor models through continual pre-training of Qwen1.5~\citep{qwen} on 140B high-quality SEA tokens and 60B tokens for replay (see Section \ref{sec:data_composition}).

\subsection{Training Infra}

\paragraph{Codebase}
To balance the training efficiency and debugging convenience, we leverage two codebases for different size model. For relatively large models (i.e., 4B and 7B), we utilize Megatron-LM~\citep{Shoeybi2019MegatronLMTM}, which supports tensor parallel and pipeline parallel to maximize the model flops utilization (MFU) of NVIDIA GPUs.
But the original Megatron codebase is a bit tricky to hands on due the absence of documents.
Thus, in practice, we employ the Megatron-LLM codebase~\footnote{\url{https://github.com/epfLLM/Megatron-LLM}}. 
It is an optimized Megatron codebase, paired with detailed documentation~\footnote{\url{https://epfllm.github.io/Megatron-LLM/}} and one-stop scripts (e.g., model sharding, data preprocessing).
For relatively small models (i.e., 0.5B and 1.8B), we employ TinyLlama~\citep{tinyllama2024} codebase~\footnote{\url{https://github.com/jzhang38/TinyLlama}}.
The codebase follows a compact and well-organised structure, which allows easy modifications for diverse purposes. 
Moreover, its optimisation enhancements significantly boost GPU utilisation. 
The combination of swift prototyping and efficient training make TinyLlama as a valuable tool in both early develop stage and final continual training stage.

\paragraph{Hardware} 
For training devices, we use NVIDIA A100 SXM4 40GB GPUs. 
To accelerate multi-node training, we further employ the InfiniBand for low latency and extreme throughput. 
During the training, we employ 
64 GPU cards for 7B\,/\,4B models, and 32 GPU cards for 1.8B\,/\,0.5B model.

\subsection{Training Details}

We adopt most of the pre-training settings and model architectures from Qwen1.5~\citep{qwen}.
It follows the standard transformer architecture~\citep{transformer}, adopts the pre-normalization with RMSNorm~\citep{Jiang2023PreRMSNormAP}, SwiGLU activation~\citep{shazeer2020glu} and rotary positional embeddings~\citep{su2022roformer}.
Notably, Qwen1.5 adds a bias item in attention of the QKV layer to improve the extrapolation ability.
Meanwhile, for the 0.5B model, we set \textsc{tie\_word\_embeddings} to False, i.e., not tying the learning of the input embedding (\textsc{embedding} module) and output projection (\textsc{lm\_head} module). Thus, the parameter of Sailor 0.5B is approximately 0.6B. However, we still name it 0.5B to be consistent with Qwen1.5.

During training, we utilize a context window length of 4,096, and integrate Flash Attention 2~\citep{Dao2023FlashAttention2FA} to improve the training efficiency and reduce the memory usage~\footnote{In contrary to Flash Attention 1~\citep{Dao2022FlashAttentionFA}, Flash Attention 2 makes it possible to train model on an arbitrary dataset that also includes padding tokens.}.
We utilize AdamW~\citep{Kingma2014AdamAM} for optimization, with the hyper-parameters $\beta_1 = 0.9, \beta_2 = 0.95, \text{eps} = 1e{-5}$.
We use the weight decay of $0.1$ and the gradient clipping of $1.0$. 
We train models with BFloat16 mixed precision to balance the training efficiency and stability.
Notably, we set \textsc{attention\_softmax\_in\_fp32} to True to execute attention masking and Softmax operations in fp32, thereby preventing precision underflow~\footnote{\url{https://github.com/huggingface/transformers/pull/17437}}.

The final pre-training corpus, SailCraft, is composed of approximately 200B tokens, integrating both SEA tokens and replay tokens, as elaborated in Section \ref{sec:data_composition}.
We use the batch size of 4M tokens and the learning rate of 1e-4.
Following a warmup period of 500 steps, the learning rate remains constant.
This scheduling strategy encourages more transferable conclusions from simulations and allows for easier recovery from interrupted training sessions.
Generally Sailor models consume around 200B tokens, completing a full pass through the SailCraft corpus once. However, the Sailor-0.5B model undergoes training with 400B tokens, equivalent to 2 epochs.

\section{Experiments}

Sailor models are evaluated on several high-quality benchmarks, including question answering, commonsense reasoning, reading comprehension and examination.

\subsection{Benchmark}

\paragraph{Question Answering} The XQuAD dataset~\citep{xquad2020} (Thai, Vietnamese) and the TydiQA dataset~\citep{tydiqa2020} (Indonesian) were selected as the representative benchmarks for question answering. The XQuAD dataset comprises 1,190 question-answer pairs from professional translations of the development set of SQuAD v1.1~\citep{squad16}. The TydiQA dataset covers 204,000 question-answer pairs directly sourced from data in their original languages, with human-written questions.

\begin{tcolorbox}[title=An example from the TydiQA dataset (Indonesian).]
\textcolor{gray}{[Context]} Mencakupi sekitar 20\% permukaan Bumi, Samudra Atlantik berada di urutan kedua terbesar dalam segi ukurannya setelah Samudra Pasifik. Bersama dengan lautan di sekitarnya ia mempunyai luas sebesar 106.450.000km²; jika lautan di sekitarnya tidak dihitung, luasnya 82.362.000km². Jumlah wilayah yang mengalir ke Samudra Atlantik lebih ... \\
\textcolor{gray}{[Qustion]} seberapa luaskah samudera atlantik? \\
\textcolor{gray}{[Answer]} 82.362.000km²
\end{tcolorbox}

\paragraph{Commonsense Reasoning} The XCOPA dataset~\citep{xcopa2020} (Indonesian, Thai, Vietnamese) provides two choices for each premise, requiring the model to select one that better addresses either the cause or effect of the event mentioned in the premise.
\begin{tcolorbox}[title=An example from the XCOPA dataset (Indonesian).]
\textcolor{gray}{[Premise]} Kunci tersebut hilang dari saku celana saya.\\
\textcolor{gray}{[Qustion]} Penyebab \\
\textcolor{gray}{[Option 1]} Saku celana saya sobek. \\
\textcolor{gray}{[Option 2]} Celana saya baru. \\
\textcolor{gray}{[Answer]} Saku celana saya sobek.
\end{tcolorbox}

\paragraph{Reading Comprehension} The BELEBELE dataset~\citep{belebele2023} is a large-scale multiple-choice machine reading comprehension benchmark spanning 122 languages. The Indonesian, Thai, and Vietnamese subsets were selected to evaluate model performance. Each question is provided with a short paragraph of context and four possible options. 
\begin{tcolorbox}[title=An example from the BELEBELE dataset (Indonesian).]
\textcolor{gray}{[Context]} Tumbuhan menghasilkan oksigen yang dihirup manusia, dan mereka menghirup karbondioksida yang dikeluarkan manusia (yang artinya, bernapas). Tumbuhan membuat makanan mereka dari matahari melalui fotosintesis. Mereka juga memberikan tempat berteduh. Kita membuat rumah dari tanaman dan membuat pakaian dari tanaman. Kebanyakan makanan yang kita makan ialah tumbuhan. Tanpa tumbuhan, hewan tidak bisa bertahan hidup. \\
\textcolor{gray}{[Qustion]} Apa yang membantu tanaman dalam proses fotosintesis? \\
\textcolor{gray}{[Option 1]} Tempat berteduh \\
\textcolor{gray}{[Option 2]} Hewan \\
\textcolor{gray}{[Option 3]} Makanan \\
\textcolor{gray}{[Option 4]} Matahari \\
\textcolor{gray}{[Answer]} Matahari
\end{tcolorbox}

\paragraph{Examination} The M3Exam dataset~\citep{m3exam2023} (Javanese, Thai, Vietnamese) is a multilingual exam benchmark collected from official school tests used in nine countries.
Note that we chose its Javanese subset since the Indonesian version has yet to be released.
\begin{tcolorbox}[title=An example from the M3Exam dataset (Javanese).]
\textcolor{gray}{[Qustion]} Pak Untung iku sabendinane lunga menyang sanggar saperlu mimpin lan ngatur lumakune crita drama \\
Miturut wacan ing inggil, pendamelan (penggawean) pak Untung dados \\
\textcolor{gray}{[Option 1]} Aktor \\
\textcolor{gray}{[Option 2]} Penulis \\
\textcolor{gray}{[Option 3]} Pelawak \\
\textcolor{gray}{[Option 4]} Sutradara \\
\textcolor{gray}{[Answer]} Sutradara
\end{tcolorbox}

\subsection{Evaluation Protocol}

As for evaluation, following established evaluation protocols, we employed the awesome evaluation platform OpenCompass~\citep{2023opencompass} to build up our evaluation code~\footnote{The code can be found at \url{https://github.com/sail-sg/sailor-llm}.}.
The performance of all models is assessed based on the 3-shot Exact Match (EM) and F1 performance, with prompts provided in native languages (e.g., Indonesian task description for Indonesian tasks).
Note that we keep the tokenizer consistent when computing the F1 scores of different models.

Following the evaluation approaches adopted in OpenCompass~\citep{2023opencompass} and the Eleuther AI evaluation framework~\citep{eval-harness} on the popular HellaSwag benchmark~\cite{hellaswag19}, we reformulated the tasks with limited output spaces (i.e., XCOPA, BELEBELE) as continuation writing tasks.
It is to say, each possible answer is appended to the given input or question, and the one that achieves the lowest perplexity score is considered as the model prediction.
As for the M3Exam dataset, we adopt the official evaluation method used in~\citet{m3exam2023} to evaluate all models. The evaluation approach involves directly prompting LLMs to produce the correct option ID when presented with a question and its corresponding options.

\subsection{Baseline Setup}

We compare Sailor models with SeaLLM~\citep{seallms2023}, Sea-Lion~\citep{SeaLion2023}, Typhoon~\citep{Typhoon2024}, and VinaLLaMA~\citep{VinaLlama2024}.
Our reporting strictly adheres to the same evaluation methodology to ensure a fair comparison, and we make much effort to closely match the reported results of all baselines.

\begin{table}[t]
\centering
\begin{tabular}{lccc}
\toprule
\textbf{3-shot (EM / F1)} &\textbf{ XQuAD (th)} & \textbf{TydiQA (id)} & \textbf{XQuAD (vi)} \\
\midrule
Qwen1.5-0.5B & 14.19 / 23.35 & 20.71 / 32.64 & 19.85 / 35.38 \\
Sailor-0.5B & 15.84 / 27.58 & 30.44 / 54.74 & 21.13 / 40.57 \\
\midrule
Qwen1.5-1.8B & 27.24 / 43.56 & 29.73 / 53.76 & 29.17 / 48.15 \\
Sailor-1.8B & 32.72 / 48.66 & 40.88 / 65.37 & 34.22 / 53.35 \\
\midrule
Qwen1.5-4B & 34.03 / 53.40 & 48.32 / 72.68 & 43.71 / 63.86 \\
Sailor-4B & 46.82 / 63.34 & 53.98 / 73.48 & 47.65 / 67.09 \\
\midrule
Llama-2-7B & 30.64 / 43.80 & 56.64 / 72.14 & 46.96 / 66.16 \\
Mistral-7B-v0.1 & 48.48 / 63.27 & 63.54 / 78.73 & 53.72 / 72.75 \\
Typhoon-7B & 51.70 / 68.92 & -- & -- \\
VinaLLaMA-7B & -- & -- & 44.82 / 64.81 \\
Sea-Lion-7B & 43.52 / 59.75 & 50.09 / 67.72 & 42.43 / 61.17 \\
SeaLLM-7B-Hybrid & 49.70 / 67.62 & 50.62 / 75.21 & 49.62 / 70.74 \\
SeaLLM-7B-v2 & 34.55 / 55.13 & 52.21 / 77.00 & 46.19 / 72.11 \\
Qwen1.5-7B & 53.79 / 69.30 & 57.17 / 77.28 & 56.63 / 76.99 \\
Sailor-7B & 57.88 / 71.06 & 60.53 / 75.42 & 53.81 / 74.62 \\
\bottomrule
\end{tabular}
\caption{Experimental results of different models on the question answering task. Note that SeaLLM-7b-Hybrid and SeaLLM-7B-v2 are both models trained with instruction tuning datasets, and the same for other tables.}
\label{tab:question_answering}
\end{table}

\begin{table}[t]
\centering
\begin{tabular}{lccc}
\toprule
\textbf{3-shot (EM)} & \textbf{XCOPA (th)} & \textbf{XCOPA (id)} & \textbf{XCOPA (vi)} \\
\midrule
Qwen1.5-0.5B & 51.00 & 52.20 & 53.80 \\
Sailor-0.5B & 51.00 & 58.20 & 58.00 \\
\midrule
Qwen1.5-1.8B & 52.60 & 51.60 & 53.40 \\
Sailor-1.8B & 53.80 & 64.20 & 63.20 \\
\midrule
Qwen1.5-4B & 53.40 & 55.00 & 57.80 \\
Sailor-4B & 53.40 & 69.20 & 68.20 \\
\midrule
Llama-2-7B & 52.80 & 64.00 & 62.00 \\
Mistral-7B-v0.1 & 57.20 & 62.40 & 61.60 \\
Typhoon-7B & 55.40 & -- & -- \\
VinaLLaMA-7B & -- & -- & 68.20 \\
Sea-Lion-7B & 60.80 & 60.60 & 67.80 \\
SeaLLM-7B-Hybrid & 58.20 & 71.60 & 67.60 \\
SeaLLM-7B-v2 & 56.80 & 64.00 & 64.60 \\
Qwen1.5-7B & 54.20 & 62.20 & 66.20 \\
Sailor-7B & 59.00 & 72.20 & 72.20 \\
\bottomrule
\end{tabular}
\caption{Experimental results of different models on the XCOPA dataset.}
\label{tab:xcopa}
\end{table}

\subsection{Experimental Results}

Experimental results shown in Table~\ref{tab:question_answering}, \ref{tab:xcopa}, \ref{tab:belebele} indicate that our Sailor models typically outperform the baseline model, Qwen1.5, in terms of performance on SEA languages.
Additionally, the performance of Sailor models is either superior or comparable to major SEA LLMs such as SeaLLM, Sea-Lion, Typhoon, and VinaLLaMA on these benchmarks.

\begin{table}[t]
\centering
\begin{tabular}{lccc}
\toprule
\textbf{3-shot (EM)} & \textbf{Belebele (th)} & \textbf{Belebele (id)} & \textbf{Belebele (vi)} \\
\midrule
Qwen1.5-0.5B & 29.89 & 26.89 & 30.22 \\
Sailor-0.5B & 32.22 & 30.89 & 32.33 \\
\midrule
Qwen1.5-1.8B & 30.11 & 32.00 & 31.33 \\
Sailor-1.8B & 34.22 & 34.89 & 35.33 \\
\midrule
Qwen1.5-4B & 32.78 & 36.22 & 35.22 \\
Sailor-4B & 36.11 & 41.33 & 38.89 \\
\midrule
Llama-2-7B & 31.78 & 39.78 & 38.00 \\
Mistral-7B-v0.1 & 34.33 & 41.33 & 41.33 \\
Typhoon-7B & 36.56 & -- & -- \\
VinaLLaMA-7B & -- & -- & 39.56 \\
Sea-Lion-7B & 36.33 & 35.56 & 37.00 \\
SeaLLM-7B-Hybrid & 37.78 & 43.11 & 43.00 \\
SeaLLM-7B-v2 & 36.33 & 43.11 & 47.00 \\
Qwen1.5-7B & 38.33 & 42.00 & 42.89 \\
Sailor-7B & 41.56 & 44.33 & 45.33 \\
\bottomrule
\end{tabular}
\caption{Experimental results of different models on the BELEBELE dataset.}
\label{tab:belebele}
\end{table}

\begin{table}[t]
\centering
\begin{tabular}{lccc}
\toprule
\textbf{3-shot (EM)} & \textbf{M3Exam (th)} & \textbf{M3Exam (jv)} & \textbf{M3Exam (vi)} \\
\midrule
Qwen1.5-0.5B &  22.38 & 22.10 & 29.12 \\
Sailor-0.5B &  21.87   & 28.84 & 23.53   \\
\midrule
Qwen1.5-1.8B &  23.81 & 26.15 & 36.39 \\
Sailor-1.8B &  23.90 &  29.65 & 27.67   \\
\midrule
Qwen1.5-4B &  26.26 & 30.19 & 40.02 \\
Sailor-4B &  27.23   &  29.11 &  31.58   \\
\midrule
Llama-2-7B & 21.13 & 23.99 & 34.15 \\
Mistral-7B-v0.1 &  29.59 & 31.00  & 43.54 \\
Typhoon-7B & 36.71 & -- & -- \\
VinaLLaMA-7B &  -- & -- & 36.95\\
Sea-Lion-7B & 23.90 & 21.56 & 26.89 \\
SeaLLM-7B-Hybrid &  25.98 & 24.53 & 38.79 \\
SeaLLM-7B-v2 &  35.60 & 29.92 & 50.36  \\
Qwen1.5-7B &  35.88 & 33.15 & 51.09 \\
Sailor-7B &  38.33 &  35.85  &  51.98\\
\bottomrule
\end{tabular}
\caption{Experimental results of different models on the M3Exam dataset.}
\label{tab:M3Exam_generation}
\end{table}

However, it is not the case for M3Exam. As shown in Table~\ref{tab:M3Exam_generation}, our Sailor models exhibit no evident advantage over Qwen1.5 at the 4B parameter scale or lower, and in certain instances, it displays noticeable weaknesses.
We have observed that the discrepancy is due to a significant option bias, which leads the Sailor models to favor certain option IDs (e.g., always C) when making predictions. Interestingly, a similar phenomenon was also observed among other baseline LLMs focusing on SEA languages.
While instruction tuning could mitigate the option bias, we have chosen not to tune the Sailor models to maintain fairness and consistency in the evaluation process.
We also provide additional results evaluated using the HellaSwag protocol in Appendix~\ref{appendix:m3exam_ppl}, which is better aligned with other benchmark results.

\section{Conclusion and Future Work}

In this paper, we present the Sailor family of open language models, tailored for South-East Asian languages, which exhibit strong performance across various multilingual tasks and benchmarks, fostering advancements in multilingual language models for the SEA region. Here are some of the most important future works:

\paragraph{Document-Friendly Deduplication}
We could improve data deduplication through document-preserving deduplication to ensure the completeness of documentation. It should adhere to the following principles:
(1) If we perform deduplication at the paragraph level, it's crucial to recombine these paragraphs into a coherent document.
(2) If a duplicated paragraph is found across multiple documents, it is advisable to retain only one instance of the paragraph in a single document while removing the duplicate occurrences from the other documents.
For example, assuming document A, B, and C each contains paragraphs a1, a2, b1, b2, c1 and c2 respectively.
The paragraph a1 is a duplicate of b1, while c1 is a duplicate of b2.
Then the algorithm should filter out the duplicated paragraphs b1 and b2 from the document B, thereby preserving the integrity and completeness of both document A and C.

\paragraph{Cross-Lingual Instruction}
In the diverse and multilingual context of South-East Asia religion, it is quite common for users to communicate in various languages. 
This creates a particularly challenging scenario for chat models, which must be adept at understanding and responding to queries in multiple languages.
For example, if the user asks the chat model a question in Indonesian, such as \textit{``Draf undangan pernikahan dalam bahasa Vietnam''} (English: \textit{Wedding invitation draft in Vietnamese}), the user would expect the model to reply in Vietnamese.
Currently, in our internal evaluations, the Sailor models does not address the challenge well.
We plan to build cross-lingual instruction datasets to address the problem.

\paragraph{Code-Switching Language Generation}
Although we introduce careful code-switching techniques to improve the model performance on language mixing, there are still some natural code-switching scenarios which are challenging for our models.
It usually involves the alternation between two or more languages within a single utterance.
The natural code-switching behavior is deeply rooted in the multilingual and multicultural societies of South-East Asia, where individuals frequently navigate between multiple linguistic and cultural backgrounds.
However, for multilingual LLMs, it is still challenging to generate code-switching texts~\citep{yong-etal-2023-prompting}.

\paragraph{More South-East Asian Languages}
To broaden the impact of open language models for SEA languages, we are dedicated to expanding our coverage to include more languages from the region.
We plan to achieve the goal by gathering high-quality training corpora from all CommonCrawl snapshots and other open resources.
Moreover, we aim to explore language generalization techniques to transfer knowledge from high-resource languages to low-resource languages, thereby enhancing the capabilities of Sailor models for the underserved languages in the SEA region.

\section*{Acknowledgement}

We extend our sincere gratitude to Zhikai Huang, Joseph Wong, and Xinyi Wan for their regular maintenance of the cluster of Sea AI Lab, ensuring its stable operation and enabling our jobs to run smoothly. We are deeply thankful to Xiaosen Zheng, Fan Zhou, Zhoujun Cheng, Binyuan Hui, Junyang Lin and Terry Yin for the fruitful discussions. We appreciate HuggingFace for providing a platform for open-source models and datasets, which have been invaluable resources in building our pre-training corpus and advancing our research.

\bibliography{tmlr}
\bibliographystyle{colm2024_conference}

\newpage

\appendix

\section{Fixing the Escape Issue in MADLAD-400 Dataset}\label{appendix:fix_madlad}
The MADLAD-400 dataset presents an issue with unicode escaping, notably the excessive occurrence of ``\textbackslash \textbackslash n'', as discussed in the HuggingFace forum\footnote{\url{https://huggingface.co/datasets/allenai/MADLAD-400/discussions/2}}. This issue could potentially disrupt the in-context learning and chatbot applications, since ``\textbackslash n'' is a common delimiter between demonstrations and task inputs. Consequently, it is necessary to replace ``\textbackslash \textbackslash n'' with either ``\textbackslash n\textbackslash n'' or ``\textbackslash n'' to resolve this problem.

\begin{algorithm}[t]
\caption{Fix Escape Issue for MADLAD-400}
\begin{algorithmic}[1]
\REQUIRE $text$
\ENSURE $processed\_text$
\STATE $processed\_text \leftarrow ""$
\STATE $sentences \leftarrow text.split("\backslash n")$
\FOR{$i \leftarrow 0$ \TO $length(sentences) - 1$}
    \STATE $has\_period\_space \leftarrow ". " \in sentences[i]$
    \STATE $next\_has\_period\_space \leftarrow i < length(sentences) - 1 \textbf{ and } ". " \in sentences[i + 1]$
    \IF{$has\_period\_space$ \textbf{ or } $next\_has\_period\_space$}
        \STATE $separator \leftarrow "\backslash n\backslash n"$
    \ELSE
        \STATE $separator \leftarrow "\backslash n"$
    \ENDIF
    \STATE $processed\_text \leftarrow processed\_text + sentences[i] + separator$
\ENDFOR
\STATE $processed\_text \leftarrow processed\_text.rstrip("\backslash n")$
\RETURN $processed\_text$
\end{algorithmic}
\label{algorithm_madlad_fixing}
\end{algorithm}

We fix the problem using a simple heuristic rule as shown in Algorithm~\ref{algorithm_madlad_fixing}.
For example, the original input is ``A.\textbackslash \textbackslash nB.\underline{\textbackslash \textbackslash n}C. D.\textbackslash \textbackslash nE. F.\textbackslash \textbackslash nG.'', where the utter-cased letter like A and G, stands for individual sentence.
The fixed output would be ``A.\textbackslash nB.\underline{\textbackslash n\textbackslash n}C. D.\textbackslash n\textbackslash nE. F.\textbackslash n \textbackslash nG.''.
The challenge (the underlined parts) is to generate the ``\textbackslash n\textbackslash n'' between B. and C.
The rules for splitting text are defined as: (1) When a paraphrase contains more than two periods, it is separated from its neighboring element with two spaces. (2) In all other cases, a single space is used as the delimiter. 
For a clearer comparison, please refer to the following specific examples.

\begin{tcolorbox}[title=Original Example with Escape Issue from the MADLAD-400 Indonesian Subset.]
Masih Diributkan, Pakar Analisis Ucapan Puan: Maknanya Dalam\underline{\textbackslash \textbackslash n}Senin, 07 September 2020 07:16 WIB\underline{\textbackslash \textbackslash n}Pakar komunikasi politik Universitas Pelita Harapan Emrus Sihombing sangat menyayangkan terjadinya penggiringan wacana negatif di ruang publik, terkait pernyataan Ketua DPP PDIP, Puan Maharani, baru-baru ini.\underline{\textbackslash \textbackslash n}Emrus mengatakan, orang yang tidak setuju lebih cenderung pendapatnya bernuansa politis dan pragmatis daripada substansi makna mendalam dari pernyataan Puan yang menyebut 'semoga Sumbar jadi pendukung negara Pancasila'.\underline{\textbackslash \textbackslash n}\"Jika kita simak dengan teori akal sehat saja, ungkapan Puan sedikitpun tidak menyebut apalagi menyinggung (perasaan) suku atau etnis tertentu yang ada di Sumbar. Diksi yang ada pada kalimat tersebut yaitu 'Sumbar' sebagai nama provinsi yaitu Sumatera Barat. Bukan suku atau etnis tertentu,\" kata Emrus, Minggu (6/9/2020).\underline{\textbackslash \textbackslash n}Baca Juga: Mbak Puan, Jangan Jadi Pemecah Bangsa\underline{\textbackslash \textbackslash n}Emrus menjelaskan, Indonesia sebagai negara kesatuan harus dimaknai bahwa setiap provinsi milik kita bersama, bukan seolah milik satu etnis atau suku tertentu, sekalipun etnis tersebut lebih dulu datang dan tinggal di provinsi tersebut dan boleh jadi lebih banyak jumlahnya.\underline{\textbackslash \textbackslash n}Warga masyarakat Sumbar, dari segi etnis atau suku sangat heterogen. Emrus menilai semua suku dari seluruh Tanah Air sudah ada di Sumbar, atau setidaknya pernah tinggal di sana. Sehingga, Sumbar bukan suku atau etnis.\underline{\textbackslash \textbackslash n}Oleh karena itu, jika ada sekelompok orang mengatasnamakan suku tertentu menolak pernyataan Puan atau berencana melaporkan ke proses hukum, tampaknya kurang pas dan bisa jadi belum melakukan pengkajian mendalam dan hilostik.\underline{\textbackslash \textbackslash n}\"Seharusnya wacana publik tertuju pada bagaimana perwujudan hak setiap individu sebagai WNI yang tinggal di Sumbar dan di semua provinsi di Indonesia dapat dijamin dan diwujudkan dalam kehidupan sehari-hari,\" ucap Emrus.\underline{\textbackslash \textbackslash n}\"Konstitusi kita, UUD 1945, menggunakan kata 'setiap' warga negara, bukan menggunakan diksi 'kelompok' atas dasar kategori sosial tertentu, termasuk etnis. Artinya, setiap individu WNI memiliki hak dan kewajiban yang sama sekalipun dari suku atau etnis yang berbeda,\" tambahnya.\underline{\textbackslash \textbackslash n}Tag: Puan Maharani, Partai Demokrasi Indonesia Perjuangan (PDIP), Pancasila
\end{tcolorbox}

\begin{tcolorbox}[title=The Corresponding Fixed Example from the MADLAD-400 Indonesian Subset.]
Masih Diributkan, Pakar Analisis Ucapan Puan: Maknanya Dalam \\
Senin, 07 September 2020 07:16 WIB \\
Pakar komunikasi politik Universitas Pelita Harapan Emrus Sihombing sangat menyayangkan terjadinya penggiringan wacana negatif di ruang publik, terkait pernyataan Ketua DPP PDIP, Puan Maharani, baru-baru ini. \\
Emrus mengatakan, orang yang tidak setuju lebih cenderung pendapatnya bernuansa politis dan pragmatis daripada substansi makna mendalam dari pernyataan Puan yang menyebut 'semoga Sumbar jadi pendukung negara Pancasila'. \\
\"Jika kita simak dengan teori akal sehat saja, ungkapan Puan sedikitpun tidak menyebut apalagi menyinggung (perasaan) suku atau etnis tertentu yang ada di Sumbar. Diksi yang ada pada kalimat tersebut yaitu 'Sumbar' sebagai nama provinsi yaitu Sumatera Barat. Bukan suku atau etnis tertentu,\" kata Emrus, Minggu (6/9/2020). \\
 \\
Baca Juga: Mbak Puan, Jangan Jadi Pemecah Bangsa \\
Emrus menjelaskan, Indonesia sebagai negara kesatuan harus dimaknai bahwa setiap provinsi milik kita bersama, bukan seolah milik satu etnis atau suku tertentu, sekalipun etnis tersebut lebih dulu datang dan tinggal di provinsi tersebut dan boleh jadi lebih banyak jumlahnya. \\
Warga masyarakat Sumbar, dari segi etnis atau suku sangat heterogen. Emrus menilai semua suku dari seluruh Tanah Air sudah ada di Sumbar, atau setidaknya pernah tinggal di sana. Sehingga, Sumbar bukan suku atau etnis. \\
 \\
Oleh karena itu, jika ada sekelompok orang mengatasnamakan suku tertentu menolak pernyataan Puan atau berencana melaporkan ke proses hukum, tampaknya kurang pas dan bisa jadi belum melakukan pengkajian mendalam dan hilostik. \\
\"Seharusnya wacana publik tertuju pada bagaimana perwujudan hak setiap individu sebagai WNI yang tinggal di Sumbar dan di semua provinsi di Indonesia dapat dijamin dan diwujudkan dalam kehidupan sehari-hari,\" ucap Emrus. \\
\"Konstitusi kita, UUD 1945, menggunakan kata 'setiap' warga negara, bukan menggunakan diksi 'kelompok' atas dasar kategori sosial tertentu, termasuk etnis. Artinya, setiap individu WNI memiliki hak dan kewajiban yang sama sekalipun dari suku atau etnis yang berbeda,\" tambahnya. \\
Tag: Puan Maharani, Partai Demokrasi Indonesia Perjuangan (PDIP), Pancasila \\
\end{tcolorbox}

\section{Data Deduplication Case Study}\label{appendix:dedup}

\begin{figure}[htb]
\centering
\includegraphics[width=1\textwidth]{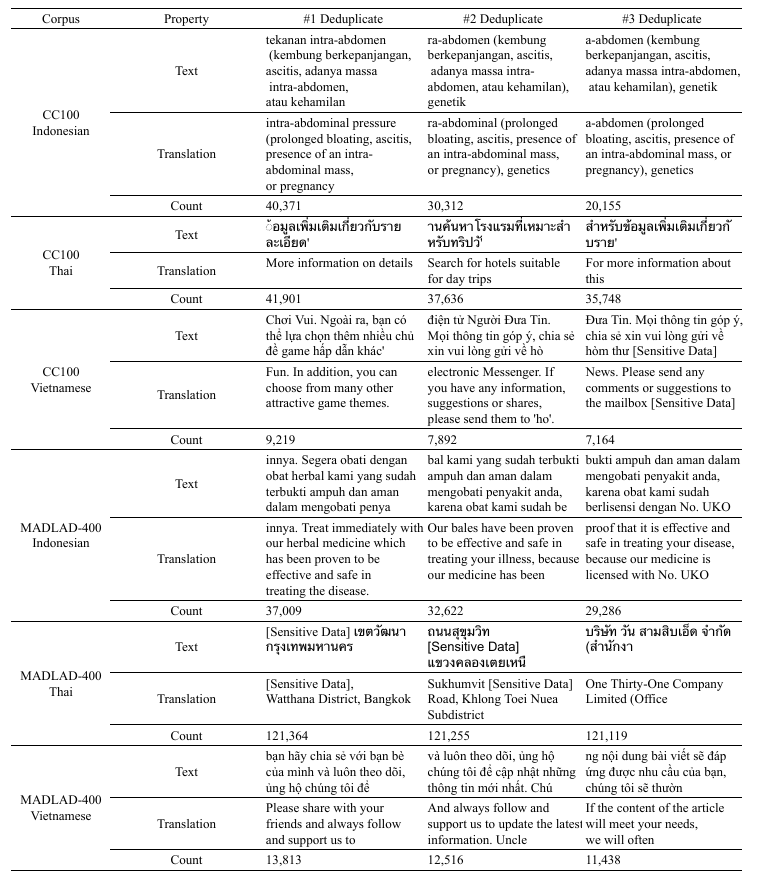}
\caption{
The top-3 frequent textual duplicates identified across CC100 and MADLAD-400 for three SEA languages, along with their respective frequencies.
Inappropriate content and personally identifiable information are replaced with \texttt{[Sensitive Data]}.
The lengthy content is truncated for clear visualization.}
\label{fig:top3_dedup_cases}
\end{figure}

We list the top-3 frequent textual duplicates for each language in Figure ~\ref{fig:top3_dedup_cases}.
To increase the distinctiveness between examples, we group them in units of 100 frequencies. Ultimately, the highest values from the first three groups are selected for demonstration.

\section{Evaluation on M3Exam by HellaSwag Style}\label{appendix:m3exam_ppl}
\begin{table}[h]
\centering
\begin{tabular}{lccc}
\toprule
\textbf{3-shot (EM)} & \textbf{M3Exam (th)} & \textbf{M3Exam (jv)} & \textbf{M3Exam (vi)} \\
\midrule
Qwen1.5-0.5B & 22.93 & 25.07 & 26.66 \\
Sailor-0.5B & 24.41 & 26.15 & 30.91 \\
\midrule
Qwen1.5-1.8B & 24.04 & 24.26 & 28.68 \\
Sailor-1.8B & 25.38 & 28.30 & 34.71 \\
\midrule
Qwen1.5-4B & 24.50 & 24.26 & 30.02 \\
Sailor-4B & 27.88 & 31.27 & 40.69 \\
\midrule
Llama-2-7B & 23.67 & 25.07 & 33.15 \\
Mistral-7B-v0.1 & 26.03 & 26.68 & 36.11 \\
Typhoon-7B & 28.53 & -- & -- \\
VinaLLaMA-7B & -- & -- & 36.22\\
Sea-Lion-7B & 25.29 & 22.91 & 38.74\\
SeaLLM-7B-Hybrid & 27.18 & 26.95 & 36.50 \\
SeaLLM-7B-v2 & 28.48 & 29.92 & 39.18 \\
Qwen1.5-7B & 25.75 & 26.15 & 36.28 \\
Sailor-7B & 30.00 & 32.88 & 44.10 \\
\bottomrule
\end{tabular}
\caption{Experimental results of different models on M3Exam using the HellaSwag style evaluation protocol.}
\label{tab:M3Exam}
\end{table}

Table~\ref{tab:M3Exam} shows the model performance on the M3Exam dataset following the evaluation approach adopted in the HellaSwag benchmark in both the Eleuther AI LM Evaluation Harness and the OpenCompass platform. We obtained the evaluation results by replacing the answer part with each possible option and appending it to the pre-defined prompt of the given question. Then, we rank the concatenated text strings by their perplexity scores provided by models and choose the one with the lowest perplexity as the model prediction.
As observed, while models evaluated using this approach generally exhibit lower performance compared to the evaluation method used in M3Exam, our inspection reveals that the models exhibit significantly reduced option bias under the evaluation protocol.

\end{document}

%% file: math_commands.tex
\usepackage{amsmath,amsfonts,bm}

\def\eqref#1{equation~\ref{#1}}

\def\1{\bm{1}}

\DeclareMathAlphabet{\mathsfit}{\encodingdefault}{\sfdefault}{m}{sl}
\SetMathAlphabet{\mathsfit}{bold}{\encodingdefault}{\sfdefault}{bx}{n}